\documentclass{article}

\usepackage{graphicx,amscd,amsmath,amssymb,verbatim}
\usepackage[dvips]{hyperref}
\usepackage{textcomp}
\usepackage[OT1,T1]{fontenc}

\usepackage{custom05e,times}

\begin{document}

\title{Stochastic Process Semantics for Dynamical Grammar Syntax:
An Overview}
\author{Eric Mjolsness \\
Department of Computer Science, and \\
Institute for Genomics and Bioinformatics \\
University of California, Irvine\\
\texttt{emj@uci.edu} \\
November 2005
}

\date{\today}

Accepted for: Ninth International Symposium on Artificial Intelligence
and Mathematics,
January 2006
\maketitle
\begin{abstract}

We define a class of probabilistic models in terms of an operator
algebra of stochastic processes, and a representation for this class
in terms of stochastic parameterized grammars.\ \ \ A syntactic
specification of a grammar is mapped to semantics given in terms
of a ring of operators, so that grammatical composition corresponds
to operator addition or multiplication.\ \ The operators are generators
for the time-evolution of stochastic processes.\ \ Within this modeling
framework one can express data clustering models, logic programs,
ordinary and stochastic differential equations, graph grammars,
and stochastic chemical reaction kinetics. This mathematical formulation
connects these apparently distant fields to one another and to mathematical
methods from quantum field theory and operator algebra.
\end{abstract}
\section{Introduction}

\ Probabilistic models of application domains are central to pattern
recognition, machine learning, and scientific modeling in various
fields.\ \ Consequently, unifying frameworks are likely to be fruitful
for one or more of these fields. There are also more technical motivations
for pursuing the unification of diverse model types. In multiscale
modeling, models of the same system at different scales can have
fundamentally different characteristics (e.g. deterministic vs.
stochastic) and yet must be placed in a single modeling framework.\ \ In
machine learning, automated search over a wide variety of model
types may be of great advantage.\ \ In this paper we propose Stochastic
Parameterized Grammars (SPG's) and their generalization to Dynamical
Grammars (DG's) as such a unifying framework.\ \ To this end we
define mathematically both the syntax and the semantics of this
formal modeling language.

\ The essential idea is that there is a ``pool'' of fully specified
parameter-bearing terms such as \{$\mathit{bacterium}( x) $, $\mathit{macrophage}(
y) $, $ \mathit{redbloodcell}( z) $\} where $x, y$ and $z$
might be position vectors. A grammar can include rules such as
\[
\left\{ \mathit{bacterium}( x) , \mathit{macrophage}( y) \right\}
\rightarrow \mathit{macrophage}( y) \; \text{\boldmath $\mathbf{with}$} \;
\rho ( \left\| x-y\right\| ) 
\]

which specify the probability per unit time, $\rho$, that the macrophage
ingests and destroys the bacterium as a function of the distance
$\|x-y\|$ between their centers.\ \ Sets of such rules are a natural
way to specify many processes.\ \ We will map such grammars to stochastic
processes in both continuous time (Section \ref{XRef-Section-92874322})
and discrete time (Section \ref{XRef-Section-9287443}), and relate
the two definitions (Section \ref{XRef-Section-92874423}).\ \ A
key feature of the semantics maps is that they are naturally defined
in terms of an algebraic {\itshape ring} of time evolution operators:
they map operator addition and multiplication into independent or
strongly dependent compositions of stochastic processes, respectively.

\ The stochastic process semantics defined here is a mathematical,
algebraic object.\ \ It is independent of any particular simulation
algorithm, though we will discuss (Section \ref{XRef-Section-92871922})
a powerful technique for generating simulation algorithms, and we
will demonstrate (Section \ref{XRef-Section-92872143}) the interpretation
of certain subclasses of SPG's as a logic programming language.\ \ Other
applications that will be demonstrated are to data clustering (\cite{StochSem05}),
chemical reaction kinetics (Section \ref{XRef-Section-92872456}),
graph grammars and string grammars (Section \ref{XRef-Section-92872523}),
systems of ordinary differential equations and systems of stochastic
differential equations (Section \ref{XRef-Section-926213036}).\ \ Other
frameworks that describe model classes that may overlap with those
described here are numerous and include: branching or birth-and-death
processes,  marked point processes, MGS modeling language using
topological cell complexes, interacting particle systems, the BLOG
probabilistic object model, adaptive mesh refinement with rewrite
rules, stochastic pi-calculus, and colored
Petri Nets.\ \ The mapping $\Psi _{c/d}$ to an
operator algebra of stochastic processes, however, appears to be
novel.

\ The present paper is an abbreviated summary of \cite{StochSem05}.

\section{Syntax Definition}

Consider the rewrite rule
\begin{equation}
A_{1}( x_{1}) ,A_{2}( x_{2}) , ...,A_{n}( x_{n}) \rightarrow B_{1}(
y_{1}) ,B_{2}( y_{2}) , ..., B_{m}( y_{m}) \;  \text{\boldmath $\mathbf{with}$} \;
\rho ( \left\{ x_{i}\right\} ,\left\{ y_{j}\right\} )  
\end{equation}

where the $A_{k}$ and $B_{l}$ denote symbols $\tau _{a}$ chosen
from an arbitrary alphabet set $\mathcal{T}=\{\tau _{a}|a\in \mathcal{A}\}$
of ``types''. In addition these type symbols carry expressions for
parameters $x_{i}$ or $y_{j}$ chosen from a base language $\mathcal{L}_{P}(
i) $ defined below. The $A$'s can appear in any order, as can the
$B$'s. Different $A$'s and $B$'s appearing in the rule can denote
the same alphabet symbol $\tau _{a}$, with equal or unequal parameter
values $x_{i}$ or $y_{j}$. $\rho $ is\ \ a nonnegative function,
assumed to be denoted by an expression in a base language $\mathcal{L}_{R}$
defined below, and also assumed to be an element of a vector space
$ \mathcal{F}$ of real-valued functions. Informally, $\rho $ is
interpreted as a nonnegative probability rate: the independent probability
per unit time that any possible instantiation of the rule will ``fire''
if its left hand side precondition remains continuously satisfied
for a small time.\ \ This interpretation will be formalized in the
semantics.\ \ 

\ We now define $\mathcal{L}_{P}( i) $. Each term $A_{i}( x_{i})
$ or $B_{j}( y_{j}) $ is of type $\tau _{a}$ and its parameters
$x_{i}$ take values in an associated (ordered) Cartesian product
set $V_{a}$ of $d_{a}$ factor spaces chosen\ \ (possibly with repetition)
from a set of base spaces $\mathcal{D}=\{D_{b}|b\in \mathcal{B}\}$.
Each $D_{b}$ is a measure space with measure $\mu _{b}$. Particular
$D_{b}$ may for example be isomorphic to the integers $ \mathbb{Z}$
with counting measure, or the real\ \ numbers $ \mathbb{R}$ with
Lebesgue measure.\ \ The ordered choice of spaces $D_{b}$ in $V_{a}=\prod
\limits_{k=1}^{d_{a}}D_{b=\sigma ( a k) } $ constitutes the type
signature $\{\sigma _{a k}\in \mathcal{B}|1\leqslant k\leqslant
d_{a}\}$ of type $\tau _{a}$. (As an aside, polymorphic argument
type signatures are supported by defining a derived type signature
$\{\sigma _{a k b}=(D_{b}\subseteq D_{\sigma ( a k) })\in \{T,F\}|1\leqslant
k\leqslant d_{a}, b\in \mathcal{B}\}$.\ \ For example we can regard
$ \mathbb{Z}$ as a subset of $ \mathbb{R}$.)\ \ Correspondingly,
parameter expressions $x_{i}$ are tuples of length $d_{a}$, such
that each component $x_{i k}$ is either a constant in the space
$D_{b=\sigma ( a k) }$, or a variable $X_{c} ( c\in \mathcal{C})$
that is restricted to taking values in that same space $D_{b( c)
}$. The variables that appear in a rule this way may be repeated
any number of times in parameter expressions $x_{i}$ or $y_{j}$
within a rule, providing only that all components $x_{i k}$ take
values in the same space $D_{b=\sigma ( a k) }$.\ \ A {\itshape
substitution} $\theta :c\mapsto D_{b( c) }$ of values for variables
$X_{c}$ assigns the same value to all appearances of each variable
$X_{c}$ within a rule.\ \ Hence each parameter expression $x_{i}$
takes values in a fixed tuple space $V_{a}$ under any substitution
$\theta$. This defines the language $\mathcal{L}_{P}( i) $.

\ We now constrain the language $\mathcal{L}_{R}$. Each nonnegative
function $\rho ( (x_{i}),(y_{j})) $ is a probability rate: the independent
probability per unit time that any particular instantiation of the
rule will fire, assuming its precondition remains continuously satisfied
for a small interval of time.\ \ It is a function only of the parameter
values denoted by $(x_{i})$ and $(y_{j})$, and not of time. Each
$\rho $ is denoted by an expression in a base language $\mathcal{L}_{R}$
that is closed under addition and multiplication and contains a
countable field of constants, dense in $ \mathbb{R}$, such as the
rationals or the algebraic numbers.\ \ $\rho$ is assumed
to be a nonnegative-valued function in a Banach space $\mathcal{F}(
V) $ of real-valued functions defined on the Cartesian product space
$V$ of all the value spaces $V_{a( i) }$ of the terms appearing
in the rule, taken in a standardized order such as nondeccreasing
order of type index $a$ on the left hand side followed by nondecreasing
order of\ \ type index $a$ on the right hand side of the rule.\ \ Provided
$\mathcal{L}_{R}$ is expressive enough, it is possible to factor
$\rho _{r}( (x_{i}),(y_{j})) $ within $\mathcal{L}_{R}$ as a product
$\rho _{r}$=$\rho _{r}^{\mathrm{pure}}( (x_{i})) $${\Pr }_{r}( (y_{j})|(x_{i}))
$ of a conditional distribution on output parameters given input
parameters ${\Pr }_{r}( (y_{j})|(x_{i})) $ and a total probability
rate $\rho _{r}^{\mathrm{pure}}( (x_{i})) $ as a function of input
parameters only.

\ With these definitions we can use a more compact notation by eliminating
the $A$'s and $B$'s, which denote types, in favor of the types themselves.
(The expression $\tau _{i}( x_{i}) $ is called a parameterized {\itshape
term,} which can match to a parameter-bearing {\itshape  object } or {\itshape
term instance }in a ``pool'' of such objects.) The caveat is that
a particular type $\tau _{i}$ may appear any finite number of times,
and indeed a particular parameterized term $\tau _{i}( x_{i}) $
may appear any finite number of times.\ \ So we use multisets ${\{...\tau
_{a( i) }( x_{i})  ... \}}_{*}$ (in which the same object $\tau
_{a( i) }( x_{i}) $ may appear as the value of several different
indices $i$) for both the LHS and RHS (Left Hand Side and Right
Hand Side) of a rule:
\begin{equation}
{\left\{ \tau _{a( i) }( x_{i}) |i\in \mathcal{I}_{L}\right\} }_{*}
\rightarrow {\left\{ \tau _{a^{\prime }( j) }( y_{j}) |j\in \mathcal{I}_{R}\right\}
}_{*} \; \text{\boldmath $\mathbf{with}$} \; \rho _{r}( \left( x_{i}\right)
,\left( y_{j}\right) ) %
\label{XRef-Equation-924145912}
\end{equation}

Here the same object $\tau _{a( i) }( x_{i}) $ may appear as the
value of several different indices $i$ under the mappings $i\mapsto
(a( i) ,x_{i})$ and/or $i\mapsto (a^{\prime }( i) ,y_{i})$.\ \ Finally
we introduce the shorthand notation $\tau _{i}=\tau _{a( i) }$ and
${\tau ^{\prime }}_{j}=\tau _{a^{\prime }( j) }$, and revert to
the standard notation $\{\}$ for multisets; then we may write $\{\tau
_{i}( x_{i}) \} \rightarrow \{{\tau ^{\prime }}_{j}( y_{j}) \}$
$\; \text{\boldmath $ \mathbf{with} $} \; \rho _{r}( (x_{i}),(y_{j})) $.

\ In addition to the {\bfseries with} clause of a rule following
the LHS$\rightarrow $RHS header, several other alternative clauses
can be used and have translations into {\bfseries with} clauses.\ \ For
example, ``{\bfseries subject to} $f( x,y) $'' is translated into
``{\bfseries with} $\delta ( f( x,y) ) $''\ \ where $\delta$ is
an appropriate Dirac or Kronecker delta function that enforces a
contraint $f( x,y) =0$. Other examples are given in \cite{StochSem05}.
The translation of ``{\bfseries solving} $e$'' or ``{\bfseries solve}
$e$'' will be defined in terms of {\bfseries with} clauses in Section
\ref{XRef-Section-926213036}.\ \ As a matter of definition, Stochastic
Parameterized Grammars do not contain {\bfseries solving}/{\bfseries
solve} clauses, but Dynamical Grammars may include them. There exists
a preliminary implementation of an interpreter for most of this
syntax in the form of a {\itshape Mathematica} notebook, which draws
samples according to the semantics of Section \ref{XRef-Section-92795243}
below.

\ A Stochastic Parameterized Grammar (SPG) $\Gamma$ consists of
(minimally) a collection of such rules with common type set $\mathcal{T}$,\ \ base
space set $ \mathcal{D}$, type signature specification $\sigma $,
and probability rate language $\mathcal{L}_{R}$.\ \ After defining
the semantics of such grammars, it will be possible to define semantically
equivalent classes of SPG's that are untyped or that have richer
argument languages $\mathcal{L}_{P}( i) $.
\section{Semantic Maps}\label{XRef-Section-92795243}

\ We provide a semantics function $\Psi _{c}( \Gamma ) $ in terms
of an operator algebra that results in a {\itshape stochastic process},
if it exists, or a special ``undefined'' element if the stochastic
process doesn't exist.\ \ The stochastic process is defined by a
very high-dimensional differential equation (the Master Equation)
for the evolution of a probability distribution in continuous time.\ \ On
the other hand we will also provide a semantics function $\Psi _{d}(
\Gamma ) $ that results in a discrete-time stochastic process for
the same grammar, in the form of an operator that evolves the probability
distribution forward by one discrete rule-firing event.\ \ In each
case the stochastic process specifies the time evolution of a probability
distribution over the contents of a ``pool'' of grounded parameterized
terms $\tau _{a}( x_{a}) $ that can each be present in the pool
with any allowed multiplicity from zero to $n_{a}^{\max }$. We will
relate these two alternative\ \ ``meanings'' of an SPG,\ \ $\Psi
_{c}( \Gamma ) $ in continuous time and $\Psi _{d}( \Gamma ) $ in
discrete time.

\ A state of the ``pool of term instances'' is defined as an integer-valued
function $n$: the ``copy number'' $n_{a}( x_{a}) \in \{0,1,2, ...\}$
of parameterized terms $\tau _{a}( x_{a}) $ that are grounded (have
no variable symbols $X_{c}$), for any combination $(a,x_{a})\in
\mathcal{V}=\coprod \limits_{a\in \mathcal{A}}a\otimes V_{a}$ of
type index $a\in \mathcal{A}$ and parameter value\ \ $x_{a}\in V_{a}$.\ \ \ We
denote this state by the ``indexed set'' notation for such functions,
$\{n_{a}( x) \}$. Each type $\tau _{a}$ may be assigned a maximum
value $n_{a}^{(\max )}$ for all $n_{a}( x_{a}) $, commonly $ \infty
$ (no constraint on copy numbers) or 1 (so $n_{a}( x_{a}) \in \{0,1\}$
which means each term-value combination is simply present or absent).\ \ The
state of the full system at time $t$ is defined as a probability
distribution on all possible values of this (already large) pool
state: $\Pr ( \{n_{a}( x_{a}) |(a,x_{a})\in \mathcal{V}\};t) \equiv
\Pr ( \{n_{a}( x_{a}) \};t) $.\ \ The probability distribution that
puts all probability density on a particular pool state $\{n_{a}(
x_{a}) \}$ is denoted $|\{n_{a}( x_{a}) \}\rangle $.

\ For continuous-time we define the semantics $\Psi _{c}( \Gamma
) $ of our grammar as the solution, if it exists, of the Master
Equation $d \Pr (  t) / d t=H\cdot  \Pr (  t) $, which can be written
out as:
\begin{equation}
\frac{d }{d t}\Pr ( \left\{ n_{a}( x) \right\} ; t) =\sum \limits_{\left\{
m_{a}( x) \right\} }H_{\left\{ n\right\}  \left\{ m\right\} } \Pr
( \left\{ m_{a}( x) \right\} ; t) %
\label{XRef-Equation-103192011}
\end{equation}

and which has the formal solution $\Pr ( t)=\exp (  t H) \cdot \Pr
( 0) $.

\ For discrete-time semantics $\Psi _{d}( \Gamma ) $ there is an
linear map $\hat{H}$ which evolves unnormalized probabilities forward
by one rule-firing time step.\ \ The probabilities must of course
be normalized, so that after $s$ discrete time steps the probability
is:
\begin{equation}
\Pr (  s) =c_{n} {\hat{H}}^{s}\cdot \Pr (  0) =\left(  {\hat{H}}^{s}\cdot
\Pr (  0) \right) /\left(  \text{\boldmath $1$}\cdot {\hat{H}}^{s}\cdot
\Pr (  0) \right) %
\label{XRef-Equation-102111134}
\end{equation}

which, taken over all $s\geqslant 0$ and $\Pr ( \{n_{a}( x) \};
0) $, defines $\Psi _{d}( \Gamma ) $.\ \ In both cases the long-time
evolution of the system may converge to a limiting distribution
$\Psi _{c}^{*}( \Gamma ) \cdot \Pr (  0) ={\lim }_{t\rightarrow
\infty }\Pr ( \{n_{a}( x) \};t) $ which is a key feature of the
semantics, but we do not define the semantics $\Psi _{c/d}( \Gamma
) $ as being only this limit even if it exists.\ \ Thus semantics-preserving
transformations of grammars are\ \ fixedpoint-preserving transformations
of grammars but the converse may not be true.

\ The Master Equation is completely determined by the {\itshape
generators} $H$ and $\hat{H}$ which in turn are simply composed
from elementary operators acting on the space of such probability
distributions.\ \ They are elements of the operator polynomial ring
$\mathbb{R}[ \{B_{\alpha }\}] $ defined over a set of basis operators
$\{B_{\alpha }\}$ in terms of operator addition, scalar multiplication,
and noncommutative operator multiplication. These basis operators
$\{B_{\alpha }\}$ provide elementary manipulations of the copy numbers
$n_{a}( x) $.
\subsection{Operator algebra}

\ The simplest basis operators $\{B_{\alpha }\}$ are elementary
creation operators $\{{\hat{a}}_{a}( x) |a\in \mathcal{A} \wedge
x\in V_{a}\}$ and annihilation operators $\{a_{a}( x) |a\in \mathcal{A}
\wedge x\in V_{a}\}$ that increase or decrease each copy number
$n_{a}( x) $ in a particular way (reviewed in \cite{MattisGlasser98}):
\begin{eqnarray}
&\left. \left. {\hat{a}}_{a}( x) |\left\{ n_{b}( y) \right\} \right\rangle
=|\left\{ n_{b}( y) +\delta _{K}( a,b) \delta _{K}( x,y) \right\}
\right\rangle  \\
&\left. \left. a_{a}( x) |\left\{ n_{b}( y) \right\} \right\rangle
=n_{a}( x) |\left\{ n_{b}( y) -\delta _{K}( a,b) \delta _{K}( x,y)
\right\} \right\rangle  
\end{eqnarray}

where $\delta _{K}( x,y) $is the Kronecker delta function. These
two operator types then generate $N_{a}( x) ={\hat{a}}_{a}( x) a_{a}(
x) $:
\[
\left. \left. \left. N_{a}( x) |\left\{ n_{b}( y) \right\} \right\rangle
={\hat{a}}_{a}( x) a_{a}( x) |\left\{ n_{b}( y) \right\} \right\rangle
=n_{a}( x) |\left\{ n_{b}( y) \right\} \right\rangle   .
\]

We can write these operators $\hat{a},a$ as finite or infinite dimensional
matrices depending on the maximum copy number $n_{a}^{(\max )}$
for type $\tau _{a}$.\ \ If $n_{a}^{(\max )}$=1 (for a fermionic
term), and we omit the type which are all assumed equal below, then
\[
\hat{a}=\left( \begin{array}{cc}
 0 & 0 \\
 1 & 0
\end{array}\right)  , a=\left( \begin{array}{cc}
 0 & 1 \\
 0 & 0
\end{array}\right) \ \ ,\ \ \hat{a}a =N\equiv \left( \begin{array}{cc}
 0  & 0  \\
 0 & 1 
\end{array}\right)  
\]

Likewise if $n_{a}^{(\max )}$=$ \infty $ (for a bosonic term), $\hat{a}=
\delta _{n,m+1}\ \ \mathrm{and}\ \ a=m \delta _{n+1,m}$. By truncating
this matrix to finite size $n^{(\max )}<\infty $ we may compute
that for some polynomial $Q( N| n^{(\max )}) $ of degree $n^{(\max
)}$-1 in $N$ with rational coefficients,
\[
\left[ a( x) , \hat{a}( y) \right] =\delta ( x-y) [ I+N Q( N| n^{\left(
\max \right) }) ] 
\]

where $\delta$ is the Dirac delta (generalized) function appropriate
to the (product) measure $\mu$ on the relevant value space $V$.
Eg. if $n^{(\max )}$=1 then $Q=-2$; if $n^{(\max )}$=$ \infty $
then $Q=0$. 
\subsection{Continuous-time semantics}\label{XRef-Section-92874322}

\ For a grammar rule number ``$r$'' of the form of (Equation \ref{XRef-Equation-924145912})
we define the operator that first (instantaneously) destroys all
parameterized terms on the LHS and then (immediately and instantaneously)
creates all parameterized terms on the RHS.\ \ This happens independently
of time or other terms in the pool.\ \ Assuming that the parameter
expressions $x, y$ contain no variables $X_{c}$, the effect of this
event is:
\begin{equation}
{\hat{O}}_{r}= \rho _{r}( \left( x_{i}\right) ,\left(
y_{j}\right) )  \left[\prod \limits_{i\in \operatorname{rhs}( r) } {\hat{a}}_{a(
i) }( x_{i}) \right] \ \ \ \left[\prod \limits_{j\in \operatorname{lhs}(
r) }a_{b( j) }( y_{j}) \right]  %
\label{XRef-Equation-922211956}
\end{equation}

If there are variables $\{X_{c}\}$, we must sum or integrate over
all their possible values in $\bigotimes \limits_{c}D_{b( c) }$:
\begin{multline}
{\hat{O}}_{r}= \int _{D_{b( 1) }}...\int _{D_{b( c) }}...\left(
\prod \limits_{c}d\mu _{b( c) }( X_{c}) \right) 
\rho _{r}( \left( x_{i}( \left\{ X_{c}\right\} ) \right)
,\left( y_{j}( \left\{ X_{c}\right\} ) \right) )
\\ 
 \left[\prod \limits_{i\in
\operatorname{rhs}( r) } {\hat{a}}_{a( i) }( x_{i}( \left\{ X_{c}\right\}
) ) \right] \ \ \ \left[\prod \limits_{j\in \operatorname{lhs}( r) }a_{b(
j) }( y_{j}( \left\{ X_{c}\right\} ) ) \right]  %
\label{XRef-Equation-922212022}
\end{multline}

Thus, syntactic variable-binding has the semantics of multiple integration.
A ``monotonic rule'' has all its LHS terms appear also on the RHS,
so that nothing is destroyed. Unfortunately ${\hat{O}}_{r}$ doesn't
conserve probability because probability inflow to new states (described
by ${\hat{O}}_{r}$) must be balanced by outflow from current state
(diagonal matrix elements).\ \ The following operator conserves
probability: $O_{r}={\hat{O}}_{r}-\operatorname{diag}( 1^{T}\cdot
{\hat{O}}_{r})  $.

For the entire grammar the time evolution operator is simply a sum
of the generators for each rule:
\begin{equation}
H=\sum \limits_{r}O_{r}=\sum \limits_{r}{\hat{O}}_{r}-\sum \limits_{r}\operatorname{diag}(
1^{T}\cdot {\tilde{O}}_{r}) \ \ =\hat{H}-D%
\label{XRef-Equation-922211931}
\end{equation}

\ This superposition implements the basic principle that every possible
rule firing is an exponential process, all happening in parallel
until a firing occurs.\ \ Note that (Equation \ref{XRef-Equation-922211956}),
(Equation \ref{XRef-Equation-922212022}) and $\hat{H}=\sum \limits_{r}{\hat{O}}_{r}$
are encompassed by the polynomial ring
$\mathbb{R}[ \{B_{\alpha }\}] $ where
the basis operators include all creation and annihilation operators.\ \ Ring
addition (as in Equation \ref{XRef-Equation-922211931} or Equation
\ref{XRef-Equation-922212022}) corresponds to independently firing
processes; ring operator\ \ multiplication (as in Equation \ref{XRef-Equation-922211956})
corresponds to obligatory event co-ocurrence of the constituent
events that define a process, in immediate succession, and nonnegative
scalar multiplication corresponds to speeding up or slowing down
a process. Commutation relations between operators describe the
exact extent to which the order of event occurrence matters.
\subsection{Discrete-time SPG semantics}\label{XRef-Section-9287443}

\ The operator $\hat{H}$ describes the flow of probability per unit
time, over an infinitesimal time interval, into new states resulting
from a single rule-firing of any type.\ \ If we condition the probability
distribution on a single rule having fired, setting aside the probability
weight for all other possibilities, the normalized distribution
is $c_{1} \hat{H}\cdot p_{0}=( \hat{H}\cdot p_{0})/( \text{\boldmath
$1$}\cdot \hat{H}\cdot p_{0})$ . Iterating, the state of the discrete-time
grammar after $s$ rule firing steps is $\Psi _{d}$ as given by (Equation
\ref{XRef-Equation-102111134}), where $\hat{H}=\sum \limits_{r}{\hat{O}}_{r}$
as before. The normalization can be state-dependent and hence dependent
on $s$, so $c_{s}\neq c^{s}$.\ \ This is a critical distinction
between stochastic grammar and Markov chain models, for which $c_{s}=c^{s}$.
An execution algorithm is directly expressed by (Equation \ref{XRef-Equation-102111134}).

\subsection{Time-ordered product expansion}\label{XRef-Section-92871922}

\ An indispensible tool for studying such stochastic processes in
physics is the time-ordered product expansion \cite{RiskenFP}.\ \ We
use the following form:
\begin{multline}
\exp ( t H) \cdot p_{0}=\exp ( t \left( H_{0}+H_{1}\right) )  \cdot
p_{0} \\
=\sum \limits_{n=0}^{\infty }\ \ \left[ \int _{0}^{t}dt_{1}\int
_{t_{1}}^{t}dt_{2}\cdots \int _{t_{n-1}}^{t}dt_{n}\exp ( \left(
t-t_{n}\right)  H_{0})  H_{1} \exp ( \left( t_{n}-t_{n-1}\right)
H_{0})  \cdots  H_{1}\exp ( t_{1} H_{0})  \right] \cdot p_{0}%
\label{XRef-Equation-92363221}
\end{multline}

where $H_{0}$ is a solvable or easily computable part of $H$, so
the exponentials $\exp ( t H_{0}) $ can be computed or sampled more
easily than $\exp ( t H) $.\ \ This expression can be used to generate
Feynman diagram expansions, in which $n$ denotes the number of interaction
vertices in a graph representing a multi-object history. If we apply
(Equation \ref{XRef-Equation-92363221}) with $H_{1}=\hat{H}$ and
$H_{0}=-D$, we derive the well-known Gillespie algorithm for simulating
chemical reaction networks \cite{Gillespie76},
which can now be applied to SPG's.
\ \ However many other decompositions
of $H$ are possible, one of which is used in Section \ref{XRef-Section-926213036}
below. Because the operators $H$ can be decomposed in many ways, there
are many valid simulation algorithms for each stochastic process.
The particular formulation 
of the time-ordered product expansion
used in (Equation \ref{XRef-Equation-92363221})
has the advantage of being recursively self-applicable. 

\ Thus, (Equation \ref{XRef-Equation-92363221})
entails a systematic approach to the creation of novel simulation algorithms.

\subsection{Relation between semantic maps}\label{XRef-Section-92874423}

{\itshape \ Proposition.} Given the stochastic parameterized grammar
(SPG) rule syntax of Equation
\ref{XRef-Equation-924145912},

\ (a) There is a semantic function $\Psi _{c}$ mapping from any
continuous-time, context sensitive, stochastic parameterized grammar
$\Gamma$ via a time evolution operator $H( \hat{H}( \Gamma ) ) $
to a joint probability density function on the parameter values
and birth/death times of grammar terms, conditioned on the total
elapsed time, $t$.

\ (b) There is a semantic function $\Psi _{d}$ mapping any discrete-time,
sequential-firing, context sensitive, stochastic parameterized grammar
$\Gamma$ via a time evolution operator $\hat{H}( \Gamma ) $ to a
joint probability density function on the parameter values and birth/death
times of grammar terms, conditioned on the total discrete time defined
as number of rule firings, $s$.

\ (c) The short-time limit of the density $\Psi _{c}( \Gamma ) $
conditioned on $t\rightarrow 0$ and conditioned on $s$ is equal
to $\Psi _{d}( \Gamma ) $.

\ Proof: (a): Section \ref{XRef-Section-92874322}. (b): Section
\ref{XRef-Section-9287443}.\ \ (c) Equation \ref{XRef-Equation-92363221}
(details in \cite{VSSTR05}, \cite{StochSem05}).

\subsection{Discussion: Transformations of SPG's}
\label{XRef-Section-emj001}

 \ Given a new kind of mathematical object (here,  SPG's or DG's) it is generally productive
in mathematics to consider the transformations of such objects
(mappings from one object to another or to itself) that preserve key properties.
Examples include transformational geometry (groups acting on lines and points)
and functors acting on categories.  In the case of SPG's, two possibilities
for the preserved property are immediately salient.  First, an SPG syntactic transformation
$\Gamma \rightarrow \Gamma^{\prime}$ could preserve the
semantics $\Psi(\Gamma)=\Psi(\Gamma^{\prime})$
either fully or just in fixed point form: $\Psi^{*}(\Gamma)=\Psi^{*}(\Gamma^{\prime})$.
Preserving the full semantics would be required of a simulation algorithm.
Alternatively,  an inference algorithm could preserve a joint probability
distribution on unobserved and observed random variables, in the form of
Bayes' rule,
$$Pr_{\Gamma}(out, internal|in) Pr(in) = Pr(in, internal, out) = Pr_{\rm Inference}(in, internal|out) Pr(out)$$
where $(in, internal, out)$ are collections of parameterized terms
that are inpuuts to, internal to, and outputs from the grammar $\Gamma$ respectively..

\section{Examples and Reductions}

\ A number of other frameworks and formalisms can be expressed or
reduced to SPGs as just defined.\ \ For example, data clustering
models are easily and flexibly described \cite{StochSem05}. We give
a sampling here.
\subsection{Biochemical reaction networks}\label{XRef-Section-92872456}

Given the chemical reaction network syntax
\begin{equation}
\left\{ m_{a}^{\left( r\right) } A_{a}|1\leqslant a\leqslant A_{\max
}\right\} \overset{k_{\left( r\right) }}{\longrightarrow }\left\{
n_{b}^{\left( r\right) }A_{b}| 1\leqslant a\leqslant A_{\max }\right\}
,%
\label{XRef-Equation-926214051}
\end{equation}

define an index mapping $a( i) =\sum \limits_{c=1}^{A_{\max }}c
\Theta (  \sum \limits_{d=1}^{c-1}m_{d}^{(r)}<i\leqslant \sum \limits_{d=1}^{c}m_{d}^{(r)})
$ and likewise for $b( j) $ as a function of $\{n_{b}^{(r)}\}$.\ \ Then
(Equation \ref{XRef-Equation-926214051}) can be translated to the
following equivalent grammar syntax for the multisets of parameterless
terms
\[
{\left\{ \tau _{a( i) }|0<i\leqslant \sum _{c=1}^{A_{\max }}m_{c}^{\left(
r\right) }\right\} }_{*} \rightarrow {\left\{ \tau _{a^{\prime }(
j) }|0<j\leqslant \sum _{c=1}^{A_{\max }}n_{c}^{\left( r\right)
}\right\}  }_{*}\ \ \ \ \ \; \text{\boldmath $\mathbf{with}$} \; k_{\left(
r\right) }
\]

whose semantics is the time-evolution generator
\begin{equation}
{\hat{O}}_{r}=  k_{\left( r\right) } \left[\prod \limits_{i\in
\operatorname{rhs}( r) } {\hat{a}}_{a( i) }\right] \ \ \ \left[\prod
\limits_{j\in \operatorname{lhs}( r) }a_{b( j) }\right] \ \ .
\end{equation}

This generator is equivalent to the stochastic process model of
mass-action kinetics for the chemical reaction network (Equation
\ref{XRef-Equation-926214051}).

\subsection{Logic programs}\label{XRef-Section-92872143}

\ Consider a logic program (e.g. in pure Prolog) consisting of Horn
clauses of positive literals
\[
p_{1}\wedge ...\wedge  p_{n}\Rightarrow q , n\geqslant 0 .
\]

Axioms have $n=0$. We can {\itshape translate} each such clause
into a monotonic SPG rule
\begin{equation}
p_{1},..., p_{n}\rightarrow q,p_{1},..., p_{n} %
\label{XRef-Equation-92354353}
\end{equation}

where each different literal $p_{i} \mathrm{or}\ \ q$ denotes an
unparameterized type $\tau _{a}$ with $n_{a}\in \{0,... n_{a}^{\max
}\}=\{0,1\}$ . Since there is no {\bfseries with} clause, the fule
firing rates default to $\rho =1$. The corresponding time-evolution
operator is
\begin{equation}
\hat{H}=\sum \limits_{r}{\hat{O}}_{r}=\sum \limits_{r}\ \ 
\left[ \prod \limits_{i\in \operatorname{rhs}( r) \setminus \operatorname{lhs}(
r) } {\hat{a}}_{a( i) }\right] \ \ \ \left[\prod \limits_{j\in \operatorname{lhs}(
r) }N_{b( j) }\right]  
\end{equation}

The semantics of the logic program is its least model or minimal
interpretation.\ \ It can be computed (Knaster-Tarski theorem) by
starting with no literals in the ``pool'' and repeatedly drawing
all their consequences according to the logic program.\ \ This is
equivalent to converging to a fixed point $\Psi ^{*}( \Gamma ) \cdot
|\text{\boldmath $0$}\rangle $ of the grammar consisting of rules
of (Equation \ref{XRef-Equation-92354353}).

More general clauses include negative literals $\neg r$ on the LHS,
as $p_{1}\wedge ... p_{n} \wedge  \neg r_{1}\wedge ... \neg r_{m}\Rightarrow
q $, or even more general cardinality constraint atoms
 $0\leqslant
l\leqslant |Z|=
\sum_{i\in A}
\Theta ( p_{i}) \leqslant u\leqslant
\infty $$\text{}$
\cite{Remmel04}.\ \ These constraints can be expressed
in operator algebra by expanding the basis operator set $\{B_{\alpha
}\}$ beyond the basic creation and annihilation operators \cite{StochSem05}.
Finally, atoms with function symbols may be admitted using parameterized
terms $\tau _{a}( x) $.
\subsection{Graph grammars}\label{XRef-Section-92872523}

\ Graph grammars are composed of local rewrite rules for graphs
(see for example \cite{GraphGram94}).\ \ We now express a class
of graph grammars in terms of SPG's. The following syntax introduces
Object Identifier (OID) labels $L_{i}$ for each parameterized term,
and allows labelled terms to point to one another through a graph
of such labels .\ \ The graph is related to two subgraphs of neighborhood
indices $N( i,\sigma ) $ and $N^{\prime }( j, \sigma ) $ specific
to the input and output sides of a rule.\ \ Like types or variables,
the label symbols appearing in a rule are chosen from an alphabet
$\{L_{\lambda }|\lambda \in \Lambda \}$.\ \ Unlike types but like
variables $X_{c}$, the label symbols $L_{\lambda ( i) }$actually
denote nonnegative integer values - unique addresses or object identifiers.

\ A graph grammar rule is of the form, for some nonnegative-integer-valued
functions $\lambda ( i) $ , $\lambda ^{\prime }( j) $, $N( i,\sigma
) $, $N^{\prime }( j,\sigma ) $ for which $(\lambda ( i) =\lambda
( j) )\Rightarrow (i=j)$,\ \ $(\lambda ^{\prime }( i) =\lambda ^{\prime
}( j) )\Rightarrow (i=j)$:
\begin{multline}
\left\{ L_{\lambda ( i) }:=\tau _{i}( x_{a( i) };\left( L_{N( i,\sigma
) }|\sigma \in 1..\sigma _{a( i) }^{\max }\right) ) |i\in \mathcal{I}\right\}
\rightarrow \left\{ L_{\lambda ( i) }|i\in \mathcal{I}_{1}\subseteq
\mathcal{I}\right\}
\\ 
 \cup \left\{ L_{\lambda ^{\prime }( j) }:=\tau
_{j}( x_{a^{\prime }( j) }^{\prime };\left( L_{N^{\prime }( j,\sigma
) }|\sigma \in 1..\sigma _{a^{\prime }( j) }^{\max }\right) ) |j\in
\mathcal{J}\right\}
\; \text{\boldmath $\mathbf{with}$} \; \rho _{r}( \left\{ x_{a^{\prime
}( j) }^{\prime }\right\} |\left\{ x_{a( i) }\right\} ) 
\end{multline}

(compare to (Equation \ref{XRef-Equation-924145912}) ).\ \ \ Note
that the fanout of the graph is limited by $\sigma _{i}^{\mathrm{cur}}\leqslant
\sigma _{a( i) }^{\max }$.\ \ Let $\mathcal{I}_{1}\mathrm{and} \mathcal{I}_{2}$
be mutually exclusive and exhaustive, and the same for $\mathcal{J}_{1}\mathrm{and}
\mathcal{J}_{2}$.\ \ Define $\mathcal{J}_{1}=\{j\in \mathcal{J}\wedge
(\exists i\in \mathcal{I}_{2}|\lambda ( i) =\lambda ^{\prime }(
j) \}$, $\mathcal{J}_{2}=\{j\in \mathcal{J}\wedge ({\nexists}i\in
\mathcal{I}_{2}|\lambda ( i) =\lambda ^{\prime }( j) \}$, and $\mathcal{I}_{3}=\{i\in
\mathcal{I}_{2}\wedge ({\nexists}j\in \mathcal{J}_{1}|\lambda (
i) =\lambda ^{\prime }( j) \}\subseteq \mathcal{I}_{2})$. Then the
graph syntax may be translated to the following ordinary non-graph
grammar rule (where NextOID is a variable, and OIDGen and Null are
types reserved for the translation):
\begin{multline*}
\left\{ \tau _{a( i) }( L_{\lambda ( i) },x_{a( i) },\left( L_{N(
i,\sigma ) }|\sigma \in 1..\sigma _{i}^{\mathrm{cur}}\right) ) |i\in
\mathcal{I}\right\}  ,\operatorname{OIDGen}( \mathrm{NextOID}) \\
\rightarrow \left\{ \tau _{a( i) }( L_{\lambda ( i) },x_{a( i) },\left(
L_{N( i,\sigma ) }|\sigma \in 1..\sigma _{i}^{\mathrm{cur}}\right)
) |i\in \mathcal{I}_{1}\right\} 
\\ 
 \cup 
\left\{ \tau _{a^{\prime }( j) }( L_{\lambda ^{\prime }( j) },x_{a^{\prime
}( j) }^{\prime },\left( L_{N^{\prime }( j,\sigma ) }|\sigma \in
1..\sigma _{j}^{\mathrm{cur}}\right) ) |
j\in \mathcal{J}_{1}\wedge \left( i\in \mathcal{I}_{2}\right)
\wedge \left( \lambda ( i) =\lambda ^{\prime }( j) \right) \right\}
\\ 
\cup 
\left\{ \tau _{a^{\prime }( j) }( L_{\lambda ^{\prime }( j) },x_{a^{\prime
}( j) }^{\prime },\left( L_{N^{\prime }( j,\sigma ) }|\sigma \in
1..\sigma _{j}^{\mathrm{cur}}\right) ) |j\in \mathcal{J}_{2}\right\}
\\ 
\cup \left\{ \operatorname{Null}( L_{\lambda ( i) }) |i\in \mathcal{I}_{3}\right\}
\cup \left\{ \operatorname{OIDGen}( \mathrm{NextOID}+|\mathcal{J}|)
\right\} \\
\; \text{\boldmath $\mathbf{with}$} \;  \rho _{r}( \left\{ x_{a^{\prime
}( j) }^{\prime }\right\} |\left\{ x_{a( i) }\right\} )  \prod \limits_{j\in
\mathcal{J}_{2}}\delta _{K}( L_{\lambda ^{\prime }( j) } , \mathrm{NextOID}+j-1)
\end{multline*}

which already has a defined semantics $\Psi _{c/d}$.\ \ Note that
all set membership tests can be done at translation time because
they do not use information that is only available dynamically during
the grammar evolution. Optionally we may also add a rule schema
(one rule per type, $\tau _{a}$) to eliminate any dangling pointers
\cite{StochSem05}.

\ Strings may be encoded as one-dimensional graphs using either
a singly or doubly linked list data structure.\ \ String rewrite
rules are emulated as graph rewrite rules, whose semantics are defined
above.\ \ This form is capable of handling many L-system grammars
\cite{PrusinkiewiczAlgB}. 

\subsection{Stochastic and ordinary differential equations}\label{XRef-Section-926213036}

\ There are SPG rule forms corresponding to stochastic differential
equations governing diffusion and transport.\ \ Given the SDE or
equivalent Langevin equation (which specializes to a system of ordinary
differential equations when $\eta ( t) =0$ ):
\begin{gather}
d x_{i}=v_{i}( \left\{ x_{k}\right\} ) d t+\sigma ( \left\{ x_{k}\right\}
) d W\ \ \ \mathrm{or}%
\label{XRef-Equation-928878}
\\\frac{d x_{i}}{d t}=v_{i}( \left\{ x_{k}\right\} ) +\eta _{i}(
t) %
\label{XRef-Equation-9288735}
\end{gather}

under some conditions on the noise term $\eta ( t) $ the dynamics
can be expressed \cite{RiskenFP} as a Fokker-Planck equation for
the probability distribution $P( \{x\},t) $:
\begin{equation}
\frac{\partial P( \left\{ x\right\} ,t) }{\partial t}=-\sum \limits_{i}\frac{\partial
}{\partial x_{i}}v_{i}( \left\{ x\right\} )  P( \left\{ x\right\}
,t) +\sum \limits_{i}\frac{\partial ^{2}}{\partial x_{i}\partial
x_{j}}D_{i j}( \left\{ x\right\} )  P( \left\{ x\right\} ,t) 
\end{equation}

Let $P( \{y\},t|\{x\},0) $ be the solution of this equation given
initial condition $P( \{y\},0) =\delta ( \{y\}-\{x\}) =\prod \limits_{k}\delta
( y_{k}-x_{k}) $ (with Dirac delta function appropriate to the particular
measure $\mu$ used for each component).\ \ Then at $t=0$,
\begin{multline*}
\frac{\partial P( \left\{ y\right\} ,0|\left\{ x\right\} ,0) }{\partial
t}\equiv \rho ( \left\{ y_{i}\right\} |\left\{ x_{i}\right\} )
  = 
-\sum \limits_{i}\frac{\partial }{\partial y_{i}}v_{i}( \left\{
x\right\} )  \delta ( \left\{ y\right\} -\left\{ x\right\} ) +\sum
\limits_{i}\frac{\partial ^{2}}{\partial y_{i}\partial y_{j}}D_{i
j}( \left\{ x\right\} )  \delta ( \left\{ y\right\} -\left\{ x\right\}
) 
\end{multline*}

Thus the probability rate $\rho ( \{y_{i}\}|\{x_{i}\}) $ is given
by a differential operator acting on a Dirac delta function. By
(Equation \ref{XRef-Equation-922212022}) we construct the evolution
generator operators $O_{\mathrm{FP}}=O_{\mathrm{drift}}+O_{\mathrm{diffusion}}$,
where
\begin{gather*}
O_{\mathrm{drift}}=-\int d\left\{ x\right\} \int d\left\{ y\right\}
\hat{a}( \left\{ y\right\} ) a( \left\{ x\right\} ) \left( \sum
\limits_{i}\nabla _{ y_{i}}v_{i}( \left\{ y\right\} ) \prod \limits_{k}\delta
( y_{k}-x_{k})  \right) 
\\O_{\mathrm{diffusion}}=\int d\left\{ x\right\} \int d\left\{ y\right\}
\hat{a}( \left\{ y\right\} ) a( \left\{ x\right\} ) \left( \sum
\limits_{i j}\nabla _{ y_{i}}\nabla _{ y_{j}}D_{i j}( \left\{ y\right\}
) \prod \limits_{k}\delta ( y_{k}-x_{k})  \right) 
\end{gather*}

The second order derivative terms give diffusion dynamics and also
regularize and promote continuity of probability in parameter
space both along and transverse to any local drift direction.
Calculations with such expressions are shown in \cite{StochSem05}.

\ Diffusion/drift rules can be combined with chemical reaction rules
to describe reaction-diffusion systems \cite{MattisGlasser98}.\ \ The
foregoing approach can be generalized to encompass partial differential
equations and stochastic partial differential equations\cite{StochSem05}.

\ These operator expressions all correspond to natural extended-time
processes given by the evolution of continuous differential equations.\ \ The
operator semantics of the differential equations is given in terms
of derivatives of delta functions.\ \ A special ``{\bfseries solve}''
or ``{\bfseries solving}'' keyword may be used to introduce such
ODE/SDE rule clauses in the SPG syntax.\ \ This syntax can be eliminated
in favor of a ``{\bfseries with}'' clause by using derivatives of
delta functions in the rate expression $\rho _{\mathrm{DE}}( \{y_{i}\}|\{x_{i}\})
$, provided that such generalized functions are in the Banach space
$\mathcal{F}( V) $ as a limit of functions.\ \ If a grammar includes
such DE rules along with non-DE rules, a solver can be used to compute
$\exp ( (t_{n+1}-t_{n}) O_{\mathrm{FP}}) $ in the time-ordered product
for $\exp ( t H) $ as a hybrid simulation algorithm for discontinuous
(jump) stochastic processes combined with stochastic differential
equations.\ \

\subsection{Discussion: Relevance to artificial intelligence and computational science}

\ The relevance of the modeling language defined here to {\it artificial intelligence}
includes the following points.  First, pattern recognition and machine
learning both benefit foundationally from better, more descriptively
adequate probabilistic domain models.  As an example,
\cite{StochSem05}
exhibits hierarchical clustering data models expressed very simply in terms of
SPG's and relates them to recent  work.  Graphical models are probabilistic
domain models with a fixed structure of variables and their relationships,
by contrast with the inherently flexible variable sets and dependency structures
resulting from the execution of stochastic parameterized grammars.
Thus SPG's, unlike graphical models, are Variable-Structure Systems
(defined in \cite{VSSTR05}), and consequently they can support compositional
description of complex situations such as multiple object tracking in the
presence of cell division in biological imagery \cite{CVPRfluor05}.
Second,
the reduction of many
divergent styles of model to a common SPG syntax and operator algebra semantics
enables new possibilities for hybrid model forms.  For example one could combine
logic programming with probability distribution models, or discrete-event
stochastic and differential equation models as discussed in Section
\ref{XRef-Section-926213036}
in possibly new ways.

\ As a third point of AI relevance,
from SPG probabilistic domain models it is possible to derive {\it algorithms}
for simulation
(as in Section \ref{XRef-Section-92871922})
and inference either by hand or automatically. Of course,
inference algorithms are not as well worked out yet for SPG's as for graphical
models. SPG's have the advantage that simulation or inference algorithms could 
be expressed again in the form of SPG's, a possibility demonstrated in part by
the encoding of logic programs as SPG's. Since both model and algorithm are expressed
as SPG's, it is possible to use SPG transformations that preserve relevant
quantities
(Section \ref{XRef-Section-emj001})
as a technique for deriving such novel algorithms
or generating them automatically.  For example we have taken this approach to
rederive by hand the Gillespie simulation algorithm for chemical kinetics. This
derivation is different from the one in Section \ref{XRef-Section-92871922}.
Because SPG's encompass
graph grammars it is even possible in principle to express families of valid
SPG transformations as meta-SPG's.
All of these points apply {\it a fortiori} to Dynamical Grammars as well.

\ The relevance of the modeling language defined here to {\it computational
science} includes the following points. First, as argued previously,
multiscale models must encompass and unify heterogeneous model types such
as discrete/continuous or stochastic/deterministic dynamical models; this
unification is provided by SPG's and DG's.  Second, a representationally
adequate computerized modeling language can be of great assistance in
constructing mathematical models in science, as demonstrated for biological
regulatory network models by Cellerator \cite{Cellerator} and other cell modeling languages.
DG's extend this promise to more complex, spatiotemporally dynamic,
variable-structure system models such as occur in biological development.
Third, machine learning techniques could in principle be applied to find
simplified approximate or reduced models of emergent phenomena within
complex domain models.  In that case the forgoing AI arguments apply to
computational science applications of machine learning as well.

\ Both for artificial intelligence and computational science, future work will
be required to determine whether the prospects outlined above are both realizable
and compelling.  The present work is intended to provide a mathematical foundation
for achieving that goal.

\section{Conclusion}

\ We have established a syntax and semantics for a probabilistic
modeling language based on independent processes leading to events
linked by a shared set of objects.\ \ The semantics is based on
a polynomial ring of time-evolution operators.\ \ The syntax is
in the form of a set of rewrite rules. Stochastic Parameterized
Grammars expressed in this language can compactly encode disparate
models: generative cluster data models, biochemical networks, logic
programs, graph grammars, string rewrite grammars, and stochastic
differential equations among other others. The time-ordered product
expansion connects this framework to powerful methods from quantum
field theory and operator algebra.

\paragraph{Acknowledgements. }Useful discussions with Guy Yosiphon,
Pierre Baldi, Ashish Bhan, Michael Duff, Sergei Nikolaev, Bruce
Shapiro, Padhraic Smyth, Michael Turmon, and Max Welling are gratefully acknowledged.
The work was supported in part by a Biomedical Information Science
and Technology Initiative (BISTI) grant (number R33 GM069013) from
the National Institue of General Medical Sciences, by the National
Science Foundation's Frontiers in Biological Research (FIBR) program
award number EF-0330786, and by the Center for Cell Mimetic Space
Exploration (CMISE), a NASA University Research, Engineering and
Technology Institute (URETI), under award number \#NCC 2-1364.

\appendix

\end{document}